\pdfoutput=1

\documentclass[11pt]{article}
\usepackage[]{acl}

\pdfoutput=1
\usepackage{multirow}
\usepackage{colortbl}

\usepackage{times}
\usepackage{latexsym}
\usepackage{xspace}

\usepackage[T1]{fontenc}

\usepackage[utf8]{inputenc}

\usepackage{microtype}

\usepackage{amsmath}
\usepackage{amsfonts}
\usepackage{graphicx,subfigure}
\usepackage{ctable}
\usepackage{comment}
\usepackage{multicol}
\usepackage{multirow}
\usepackage{epsfig}
\usepackage{pifont}
\usepackage{dsfont}
\usepackage{makecell}
\usepackage{adjustbox}
\usepackage{txfonts}

\usepackage[boxed, ruled, vlined, linesnumbered]{algorithm2e}
\usepackage{amstext}

\title{Are Shortest Rationales the Best Explanations\\for Human Understanding?}

\definecolor{ao(english)}{rgb}{0.0, 0.5, 0.0}

\newcommand{\kenneth}[1]{}
\newcommand{\hua}[1]{}
\newcommand{\wenbo}[1]{}
\newcommand{\sherry}[1]{}

\newcommand{\system}{\textsc{LimitedInk}\xspace}

\newcommand{\eg}{\emph{e.g.,}\xspace}%
\newcommand{\ie}{\emph{i.e.,}\xspace}

\renewcommand{\textsuperscript}[1]{\raisebox{0.5ex}{#1}}

\author{
Hua Shen\textsuperscript{${\dagger}$}
\quad Tongshuang Wu\textsuperscript{${\diamondsuit}$}
\quad Wenbo Guo\textsuperscript{${\dagger}$}
\quad Ting-Hao `Kenneth' Huang\textsuperscript{${\dagger}$}\\ 
    \textsuperscript{${\dagger}$}College of Information Sciences and Technology, Pennsylvania State University\\
    \textsuperscript{${\diamondsuit}$}Paul G.\ Allen School of Computer Science and Engineering, University of Washington\\
    {\tt \{huashen218,wzg13,txh710\}@psu.edu}\\
    {\tt wtshuang@cs.washington.edu}\\
  }

\begin{document}

\maketitle

\begin{abstract}
Existing self-explaining models typically favor extracting the shortest possible rationales --- snippets of an input text ``responsible for'' corresponding output --- to explain the model prediction, with the assumption that shorter rationales are more intuitive to humans.
However, this assumption has yet to be validated.
Is the shortest rationale indeed the most human-understandable?
To answer this question, we design a self-explaining model, \system,
which allows users to extract rationales at any target length.
Compared to existing baselines, \system achieves compatible end-task performance and human-annotated rationale agreement, making it a suitable representation of the recent class of self-explaining models.
We use \system to conduct a user study on the impact of rationale length, 
where we ask human judges to predict the sentiment label of documents based only on \system-generated rationales with different lengths.
We show rationales that are too short do not help humans predict labels better than randomly masked text, suggesting the need for more careful design of the best human rationales.\footnote{\label{note}Find open-source code at: \url{https://github.com/huashen218/LimitedInk.git}}

\end{abstract}

\section{Introduction}
\label{sec:introduction}


\begin{figure}[!t]
    \centering
    \includegraphics[width=1.0\columnwidth]{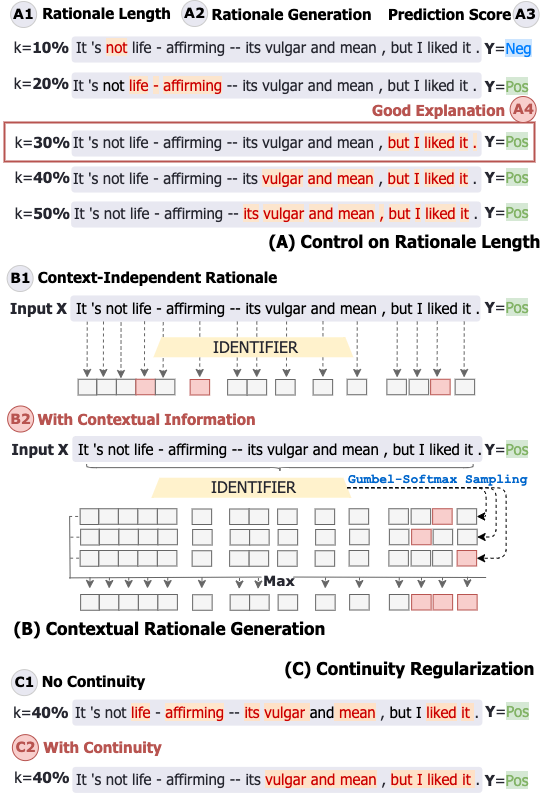}
    \vspace{-1.2pc}
    \caption{
    \system's rationale generation with length control: (A) control rationale generation with different lengths; (B) incorporating contextual information into rationale generation; (C) regularizing continuous rationale for human interpretability. Examples use the SST dataset for sentiment analysis~\cite{socher-etal-2013-recursive}.
    }
    
    \label{fig:example}
    \vspace{-.8pc}
 \end{figure}

While neural networks have recently led to large improvements in NLP, most of the models make predictions in a black-box manner, making them indecipherable and untrustworthy to human users.
In an attempt to faithfully explain model decisions to humans, various work has looked into extracting \emph{rationales} from text inputs~\cite{jain-etal-2020-learning,paranjape-etal-2020-information}, 
with \textit{rationale} defined as the ``shortest yet sufficient subset of input to predict the same label''~\cite{lei-etal-2016-rationalizing,bastings-etal-2019-interpretable}.
The underlying assumption is two-fold: 
(1) by retaining the label, we are extracting the texts used by predictors~\cite{jain-etal-2020-learning}; 
and (2) short rationales are more readable and intuitive for end-users, and thus preferred for human understanding~\cite{vafa2021rationales}.
Importantly, prior work has knowingly traded off some amount of model performance to achieve the shortest possible rationales.
For example, when using less than 50\% of text as rationales for predictions, \citet{paranjape-etal-2020-information} achieved an accuracy of 84.0\% (compared to 91.0\% if using the full text).
However, the assumption that the shortest rationales have better human interpretability has not been validated by human studies~\cite{shen2021explaining}.
Moreover, when the rationale is too short, the model has much higher chance of missing the main point in the full text.
In Figure~\ref{fig:example}$A$, although the model can make the correct positive prediction when using only 20\% of the text, it relies on a particular adjective, ``life-affirming,'' which is seemingly positive but does not reflect the author's sentiment.
These rationales may be confusing when presented to end-users.

In this work, we ask: \textit{Are shortest rationales really the best for human understanding?}
To answer the question, we first design \system, a self-explaining model that flexibly extracts rationales at any target length (Figure~\ref{fig:example}$A$).
\system allows us to control and compare rationales of varying lengths on 
input documents.
Besides \textbf{controls on rationale length}, we also design \system's sampling process and objective function to be \textbf{context-aware} (\ie rank words based on surrounding context rather than individually, Figure~\ref{fig:example}$B_2$) and \textbf{coherent} (\ie prioritize continuous phrases over discrete tokens, Figure~\ref{fig:example}$C_2$).
Compared to existing baselines (\eg \texttt{Sparse-IB}
), \system achieves compatible end-task performance and alignment with human annotations on the ERASER~\cite{deyoung2019eraser} benchmark, which means it can represent recent class of self-explaining models.

We use \system to conduct user studies to
investigate the effect of rationale length on human understanding.
%
%
Specifically, we ask MTurk participants to predict document sentiment polarities based on only \system-extracted rationales.
%
%
%
By contrasting rationales at five different length levels, we find that shortest rationales are largely not the best for human understanding.
In fact, humans do not perform better prediction accuracy and confidence better than using randomly masked texts when rationales are too short (\eg 10\% of input texts).
%
%
%
%
%
%
%
In summary, this work encourages a rethinking of self-explaining methods to find the right balance between brevity and sufficiency.
\section{\system}
\label{sec:method}
\subsection{Self-Explaining Model Definition}
\label{sec:definition}


We start by describing typical self-explaining methods~\cite{lei-etal-2016-rationalizing,bastings-etal-2019-interpretable,paranjape-etal-2020-information}.
Consider a text classification dataset containing each document input as a tuple $(\mathbf{x}, y)$. Each input $\mathbf{x}$ includes $n$ features (\emph{e.g.,} sentences or tokens) as $\mathbf{x}=[x_1,x_2,...,x_n]$, and $y$ is the prediction.
%
%
%
%
The model typically consists of an \emph{identifier} $\mathbf{idn}(\cdot)$ to derive a boolean mask $\mathbf{m}=[m_1, m_2, ..., m_n]$, where $m_i \in \{1,0\}$ indicates whether feature $x_i$ is in the rationale or not. Note that the mask $\mathbf{m}$ is typically a binary selection from the \emph{identifier}'s probability distribution, i.e., $\mathbf{m} \sim \mathbf{idn}(\mathbf{x})$.
Then it extracts rationales $\mathbf{z}$ by $\mathbf{z} =\mathbf{m} \odot \mathbf{x}$, and further leverages a \emph{classifier} $\mathbf{cls}(\cdot)$ to make a prediction $y$ based on the identified rationales as $y = \mathbf{cls}(\mathbf{z})$. The optimization objective is:
\vspace{-1.5em}
\begin{equation}
\begin{array}{l}
\displaystyle \min_{\theta_{\mathbf{idn}}, \theta_{\mathbf{cls}}} \  \underbrace{\mathbb{E}_{\mathbf{z}\sim\mathbf{idn}(\mathbf{x})} \mathcal{L}(\mathbf{cls}(\mathbf{z}),y)
}_{\small\mbox{sufficient prediction}} 
+ \underbrace{\lambda \Omega (\mathbf{m})}_{\small\mbox{regularization}} \\
\small
\label{eq:objective}
\vspace{-1.5em}
\end{array}
\end{equation}
where $\theta_{\mathbf{idn}}$ and $\theta_{\mathbf{cls}}$ are trainable parameters of \emph{identifier} and \emph{classifier}. $\Omega (\mathbf{m})$ is the regularization function on mask and $\lambda$ is the hyperparameter.


\begin{table*}[!t]
\setlength{\tabcolsep}{3.6pt}
\small
\footnotesize
\centering
\begin{tabular}{@{} r|c c c c |c c c c |c c c c |c c c c |c c c c @{}}
\toprule
\multirow{2}{*}{\textbf{Method} } & \multicolumn{4}{c|}{\cellcolor[HTML]{DAE8FC}\textbf{Movies}}  & \multicolumn{4}{c|}{\cellcolor[HTML]{DAE8FC}\textbf{BoolQ}}  & \multicolumn{4}{c|}{\cellcolor[HTML]{DAE8FC}\textbf{Evidence Inference}}  & \multicolumn{4}{c|}{\cellcolor[HTML]{DAE8FC}\textbf{MultiRC}}  & \multicolumn{4}{c}{\cellcolor[HTML]{DAE8FC}\textbf{FEVER}}  \\
& Task & P & R & F1 & Task & P & R & F1 & Task & P & R & F1 & Task & P & R & F1 & Task & P & R & F1  \\
\midrule\midrule
\texttt{Full-Text}        & .91  & - & - & -  & .47  & - & - & -  & .48  & - & - & - & .67  & - & - & - & .89  & - & - & - \\
\midrule
\texttt{Sparse-N}     & .79  & .18 & .36 & .24  & .43  & .12          &  .10 & .11  & .39  & .02 & .14 & .03  & .60 & .14  & .35 & .20  & .83  & .35 & .49 & .41 \\  
\texttt{Sparse-C}      & .82  & .17 & .36 & .23  & .44  & .15          &  .11 & .13  & .41  & .03 & .15 & .05  & .62 & .15  & \textbf{.41} & .22  & .83  & .35 & .52 & .42\\
\texttt{Sparse-IB}     & .84  & .21 & .42 & .28  & .46  & \textbf{.17} & .15  & .15  & .43  & .04 & .21 & .07  & .62 &  .20 & .33  & .25  & .85  & \textbf{.37} & .50 & \textbf{.43} \\
\midrule
\rowcolor[HTML]{EFEFEF}
\system & \textbf{.90} & \textbf{.26} & \textbf{.50} &  \textbf{.34} & \textbf{.56} & .13 & \textbf{.17} & \textbf{.15}  & \textbf{.50}  & \textbf{.04} & \textbf{.27} & \textbf{.07}  & \textbf{.67}  & \textbf{.22} & .40 & \textbf{.28}  &  \textbf{.90} &  .28 & \textbf{.67} & .39 \\
%
\cellcolor[HTML]{EFEFEF} Length Level   & \multicolumn{4}{c|}{\cellcolor[HTML]{EFEFEF} 50\% } &  \multicolumn{4}{c|}{\cellcolor[HTML]{EFEFEF} 30\% }  & \multicolumn{4}{c|}{\cellcolor[HTML]{EFEFEF} 50\%}   &  \multicolumn{4}{c|}{\cellcolor[HTML]{EFEFEF} 50\%} & \multicolumn{4}{c}{\cellcolor[HTML]{EFEFEF} 40\%}  \\
\bottomrule
\end{tabular}
\vspace{-.5pc}
\caption{
\system performs compatible with baselines in terms of end-task performance (\textbf{Task}, weighted average F1) and human annotated rationale agreement (\textbf{P}recision, \textbf{R}ecall, \textbf{F1}).
All results are on test sets and are averaged across five random seeds.
%
%
%
%
%
For \system, we report results for the best performing \emph{length level}.
}
\label{tab:prediction}
\vspace{-4mm}
\end{table*}

\vspace{-0.5em}
\subsection{Generating Length Controllable Rationales with Contextual Information}
We next elaborate on the definition and method of controlling rationale length in \system
%
%
Assuming that the rationale length is $k$ as prior knowledge, we enforce the generated boolean mask to sum up to $k$  as $k = \sum_{i=1}^{n}(m_i)$, where $\mathbf{m} = \mathbf{idn}(\mathbf{x}, k)$.
Existing self-explaining methods commonly solve this by sampling from a Bernoulli distribution over input features, thus generating each mask element $m_i$ independently conditioned on each input feature $x_i$~\cite{paranjape-etal-2020-information}.
For example, in Figure~\ref{fig:example}$B_1$), ``life affirming'' is selected independent of the negation context ``not'' before it, which contradicts with the author's intention.
However, these methods potentially neglect the contextual input information. We leverage the concrete relaxation of subset sampling
%
technique~\cite{L2X:2018:icml} to incorporate contextual information into rationale generation process (see Figure~\ref{fig:example}$B_2$), where we aim to select the top-k important features over all $n$ features in input $\mathbf{x}$ via Gumbel-Softmax Sampling (\emph{i.e.,} applying the Gumbel-softmax trick to approximate weighted subset sampling process).
%
To further guarantee precise rationale length control, we deploy the 
\emph{vector and sort} regularization on mask $\mathbf{m}$~\cite{fong:2019:iccv}.
See more model details in Appendix~\ref{appdx:models}.

\vspace{-0.5em}
\subsection{Regularizing Rationale Continuity}
\label{sec:regularizer}
To further enforce coherent rationale for human interpretability, we employ the Fused Lasso to encourage continuity property~\cite{jain-etal-2020-learning,bastings-etal-2019-interpretable}.
The final mask regularization is:
\vspace{-1.5em}
\begin{equation}
\small
\begin{array}{l}
\Omega(\mathbf{m}) = \lambda_1 \underbrace{\sum_{i=1}^{n} |m_{i}-m_{i-1}|}_{\small{\mbox{Continuity}}}
 + \lambda_2 \underbrace{\| \operatorname{vecsort}\left(m\right)-\hat{m} \|}_{\small{\mbox{Length Control}}} 
\end{array}
\label{eq:regularizer}
\vspace{-1mm}
\end{equation}

For BERT-based models, which use subword-based tokenization algorithms (\emph{e.g.,} WordPiece), we assign each token's importance score as its sub-tokens' maximum score to extract rationales during model inference (see Figure~\ref{fig:example}$C$).
\section{Model Performance Evaluation}
\label{sec:experiment}

We first validate \system on two common rationale evaluation metrics, including end-task performance and human annotation agreement.

%

\vspace{-0.5em}
\subsection{Experimental Setup}

We evaluate our model on five text classification datasets from the ERASER benchmark~\cite{deyoung2019eraser}.
We design the \textit{identifier} module in \system as a BERT-based model, followed by two linear layers with the ReLU function and dropout technique.
The temperature for Gumbel-softmax approximation is fixed at 0.1.
Also, we define the \textit{classifier} module as a BERT-based sequence classification model to predict labels.
We train five individual self-explaining models of different rationale lengths with training and validation sets, where we set the rationale lengths as $\{10\%$, $20\%$, $30\%$, $40\%$, $50\%\}$ of all input text.
Then we select one out of the five models, which has the best weighted average F1 score, to compare with current baselines on end-task performance and human annotation agreement on test sets.
Note that we use all models with five rationale lengths in human evaluation described in Section~\ref{sec:evaluation}.

\vspace{-.6em}
\paragraph{Baselines.}
We compare \system with four baselines. 
\texttt{Full-Text} consists of only the \textit{classifier} module with full-text inputs.
\texttt{Sparse-N} enforces shortest rationales by minimizing rationale mask length~\cite{lei-etal-2016-rationalizing,bastings-etal-2019-interpretable}.
\texttt{Sparse-C} controls rationale length by penalizing the mask when its length is less than a threshold~\cite{jain-etal-2020-learning}.
\texttt{Sparse-IB} enables length control by minimizing the KL-divergence between the generated mask with a prior distribution~\cite{paranjape-etal-2020-information}.
See Appendix~\ref{appdx:models} for more model and baseline details.

\vspace{-0.5em}
\subsection{Evaluation Results}

\vspace{-0.2em}
\paragraph{End-Task Performance.}
Following metrics in~\citet{deyoung2019eraser}, we report the weighted average F1 scores for end-task classification performance.
Among five \system models with different rationale lengths, Table~\ref{tab:prediction} reports the model with the best end-task performance on the test set.
%
%
%
%
We observe that \system performs similarly to or better than the self-explaining baselines in all five datasets. See ablation studies in Appendix~\ref{appdx:ablation}.

\vspace{-.6em}
\paragraph{Human-Annotated Rationale Agreement.}
We calculate
the alignment between generated rationales and human annotations collected in the ERASER benchmark~\cite{deyoung2019eraser}.
As also shown in Table~\ref{tab:prediction}, we report the Token-level F1 (F1) metric along with corresponding Precision (P) and Recall (R) scores.
The results show that \system can generate rationales that are consistent with human annotations and comparable to self-explaining baselines in all datasets.

\section{Human Evaluation}
\label{sec:evaluation}

Equipped with \system, we next carry out human studies to investigate the effect of rationale length on human understanding.
 

\vspace{-0.5em}
\subsection{Study Design}

\begin{figure}[!t]
    \centering
    \includegraphics[width=\columnwidth]{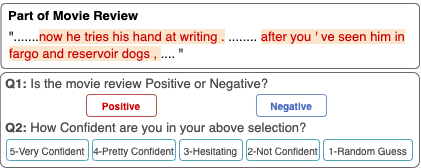}
    \vspace{-1.5pc}
    \caption{Key components of the User Interface in the MTurk \emph{task} HITs.
    Note that each HIT contains five reviews with different rationale lengths.
    }
    \label{fig:user-interface}
\end{figure}

\begin{figure}[!t]
    \centering
    \includegraphics[width=\columnwidth]{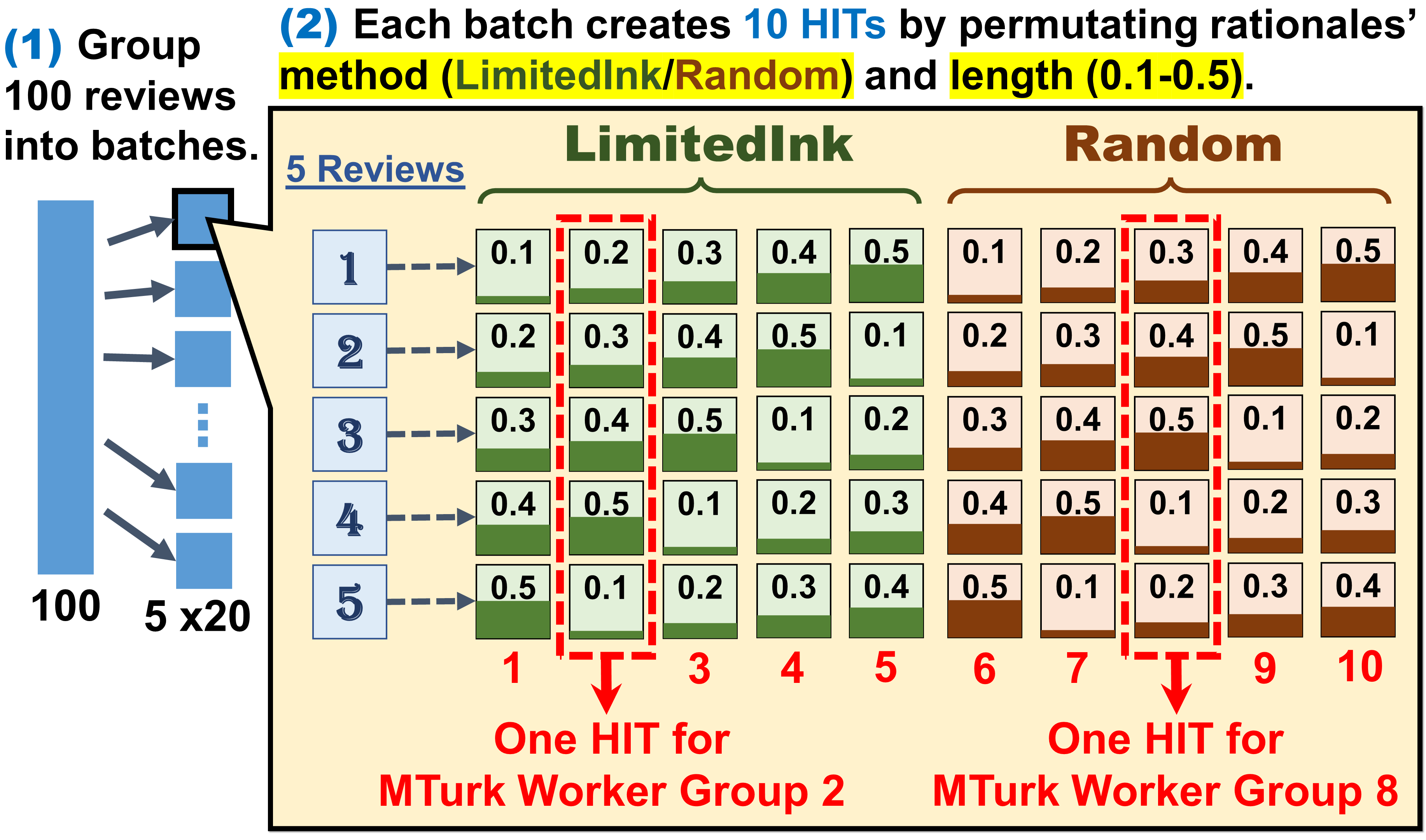}
    \vspace{-1.3pc}
    \caption{The human evaluation's workflow.
    We (1) divide 100 movie reviews into 20 batches and (2) produce 10 HITs from each batch for ten worker groups.
    }
    \label{fig:mturk-workflow}
    \vspace{-.8pc}
 \end{figure}

Our goal is to quantify human performance on predicting the labels and confidence based solely on the rationales with different lengths.
To do so, we control \system to extract rationales of different lengths, and recruit Mechanical Turk (MTurk) workers to provide predictions and confidence.
 
\vspace{-.5em}
\paragraph{Dataset \& rationale extraction.}
We focus on sentiment analysis in user study, and randomly sample 100 reviews from the Movie Reviews~\cite{zaidan-eisner-2008-modeling} test set that have correct model predictions.
Then, we extract five rationales for each review using \textbf{\system}, with lengths from 10\% to 50\%, with an increment of 10\%.

Since human accuracy likely increases when participants see more words (\ie when the lengths of rationales increase), we also create a \textbf{\texttt{Random}} rationale baseline, where we randomly select words of the same rationale length on the same documents (10\% to 50\%) while taking the continuity constraint into consideration. More details of \texttt{Random} baseline generation are in~Appendix~\ref{appdx:random_baseline}. 

\vspace{-.5em}
\paragraph{Study Procedure.}
The study is completed in two steps.
First, we posted a \emph{qualification} Human Intelligence Tasks (HITs, \$0.50 per assignment) on MTurk to recruit 200 qualified workers.\footnote{In addition to our custom qualification used for worker grouping, three built-in worker qualifications are used in all of our HITs: HIT Approval Rate ($\geq $98\%), Number of Approved HITs ($\geq 3000$), and Locale (US Only) Qualification. }
%
%
Next, the 200 recruited workers can participate the \emph{task} HIT (\$0.20 per assignment, 7 assignments posted) which contains five distinct movie reviews, with varying rationale lengths (10\%-50\%).
In \emph{task} HIT, as key components shown in Figure~\ref{fig:user-interface}, we only display the rationales and mask all other words with ellipses of random length, such that participants can not infer the actual review length.
Then participants are asked to guess the sentiment of the full review, and provide their confidence level based on a five-point Likert Scale~\cite{likert1932technique}.
The full user interface is in Appendix~\ref{appdx:interface}.

\vspace{-.5em}
\paragraph{Participants recruiting and grouping.} 
%
With each review having ten distinct rationales (five from \system and five \texttt{Random}), if these rationale conditions were randomly assigned, participants are likely to see the same review repeatedly and gradually see all the words. 
We carefully design our study to eliminate such undesired learning effect.
%
%
%
%
More specifically, we group our 100 reviews into 20 batches, with five reviews in each batch (Step 1 in Figure~\ref{fig:mturk-workflow}).
For each batch, we create five HITs for \system and \texttt{Random}, respectively, such that all the rationale lengths of five reviews are covered by these 10 HITs (Step 2 in Figure~\ref{fig:mturk-workflow}). 
Further, we make sure each participant is only assigned to one unique HIT, so that each participant can only see a review once. To do so, we randomly divide the 200 qualified workers into 10 worker groups (20 workers per group), and pair one worker group with only one HIT in each batch.
%
This way, each HIT can only be accomplished by one worker group.
As our participant control is more strict than regular data labeling tasks on MTurk, we keep the HITs open for 6 days. 
110 out of 200 distinct workers participated in the main study,
and they completed 1,169 of 1,400 assignments.

\begin{figure}[!t]
    \centering
    \includegraphics[width=1.0\columnwidth]{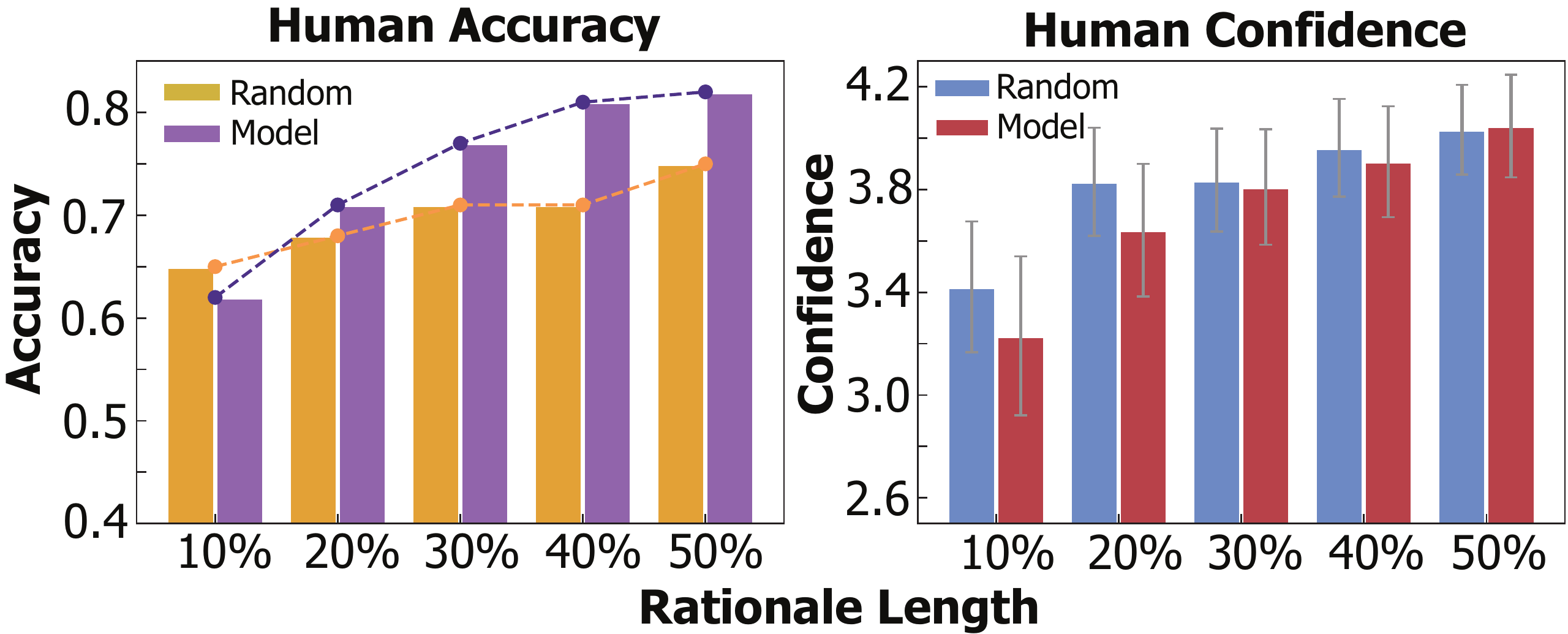}
     \vspace{-1.5pc}
    \caption{Human accuracy and confidence on predicting model labels given rationales with different lengths.
    }
    \label{fig:human}
    \vspace{-10pt}
 \end{figure}

\begin{table}[!t]
\footnotesize
\setlength{\tabcolsep}{3pt}
\begin{tabular}{@{} r | r|cc @{} }
\toprule
\multicolumn{2}{c|}{length level (\%)} 
    & \cellcolor[HTML]{DAE8FC} \textbf{Negative} 
    &  \cellcolor[HTML]{DAE8FC}\textbf{Positive} \\ 
\multicolumn{2}{c|}{ \& Extract. method} 
    &  P  /  R  /  F1 
    &  P  /  R  /  F1                \\ 
\midrule\midrule
\multirow{2}{*}{10\%} 
    &\system & 0.66 / 0.56 / / 0.61        & \textbf{0.70} / 0.58 / 0.64 \\ 
    &\texttt{Random} &\textbf{0.67 / 0.57 / 0.62} & 0.66 / \textbf{0.70 / 0.68} \\ 
\midrule
\multirow{2}{*}{20\%} 
    & \system &  \textbf{0.75 /  0.61 / 0.67}       &  \textbf{0.71 / 0.77 / 0.74}                           \\ 
   & \texttt{Random} &   0.69 / 0.60 / 0.64     &   0.68 / 0.74 / 0.71   \\ 
\midrule
\multirow{2}{*}{30\%}
    & \system & \textbf{0.74 / 0.76 / 0.75}   & \textbf{0.81 / 0.78 / 0.79}                  \\ 
    & \texttt{Random} &  0.72 / 0.61 / 0.66     &  0.72 / 0.78 / 0.75    \\ 
\midrule
\multirow{2}{*}{40\%}
    & \system & \textbf{0.84 / 0.76 / 0.80}   &      \textbf{0.78 / 0.85 / 0.81}  \\ 
    & \texttt{Random} &    0.79 / 0.63 / 0.70   &    0.65 / 0.79 / 0.71   \\ 
\midrule
\multirow{2}{*}{50\%}
    & \system & \textbf{0.78 / 0.78 / 0.78} &  \textbf{0.85 / 0.84 / 0.85}                  \\ 
    & \texttt{Random} &  0.77 / 0.63 / 0.70     &  0.75 / 0.84 / 0.79 \\ 
\bottomrule
\end{tabular}
\vspace{-.5pc}
\caption{Human performance (\emph{i.e.,} Precision / Recall / F1 Score) on predicting model labels of each category in the Movie Reviews dataset.}
\label{tab:human}
\vspace{-1.pc}
\end{table}

\vspace{-.5em}
\subsection{Results}
%
We show the human prediction accuracy and confidence results in Figure~\ref{fig:human}.
We find that the best explanations for human understanding are largely not the shortest rationales (10\% length level): here, the human accuracy in predicting model labels
is lower than for the random baseline (0.61 vs. 0.63), indicating that the shortest rationales are not the best for human understanding.
%
%
%
There is a significant difference in human predicted labels (\ie ``positive''=1,``negative''=2) between \system (M=1.24,SD=0.71) and \texttt{Random} (M=1.32,SD=0.54); t(1169)=2.27, p=0.02.
%
Table~\ref{tab:human} shows human performance for each category.

Additionally, notice that the slope of our model's accuracy consistently flattens as the rationale increases, whereas the random baseline does not display any apparent trend and is obviously lower than our model at higher length levels ({\em e.g.}, 40\%).
We hypothesize that this means our model is (1) indeed learning to reveal useful rationales (rather than just randomly displaying meaningless text), and (2) the amount of information necessary for human understanding only starts to saturate at around 40\% of the full text.
This creates a clear contrast with prior work, where most studies extract 10-30\% of the text as the rationale on the same dataset~\cite{jain-etal-2020-learning,paranjape-etal-2020-information}.
The eventually flattened slope potentially suggests a sweet spot to balance human understanding on rationales and sufficient model accuracy.

\section{Discussion}
\label{sec:discussion}


By examining human prediction performance on five levels of rationale lengths, we demonstrate that the shortest rationales are largely not the best for human understanding.
We are aware that this work has limitations. The findings are limited to Movie Reviews dataset, and we only evaluate human performance with rationales generated by the proposed \system.
Still, our findings challenge the ``shorter is better'' assumption commonly adopted in existing self-explaining methods. As a result, we encourage future work to more cautiously define the best rationales for human understanding, and trade off between model accuracy and rationale length.
More concretely, we consider that rationale models should find the right balance between brevity and sufficiency.
One promising direction could be to clearly define the optimal human interpretability in a measurable way and then learn to adaptively select rationales with appropriate length.

\section{Related Work}
\label{sec:literature}
\textbf{Self-explaining models.} 
Self-explaining models, which condition predictions on their rationales, are considered more trustworthy than post-hoc explanation techniques~\cite{rajagopal2021selfexplain}.
However, existing efforts often enforce minimal rationale length, which degrade the predictive performance~\cite{yu-etal-2019-rethinking,bastings-etal-2019-interpretable,jain-etal-2020-learning}.
\citet{paranjape-etal-2020-information} improves this by proposing an information bottleneck approach to enable rationale length control at the sentence level.
In this paper, \system further enables length control at the token level to allow more flexibility needed for our human studies.

\vspace{-.6em}
\paragraph{Human-grounded evaluation.} 
A line of studies evaluated model-generated rationales by comparing them against human-annotated explanations~\cite{carton-etal-2020-evaluating,paranjape-etal-2020-information}.
Some other studies collect feedback from users to evaluate the explanations, such as asking people to choose a preferred model~\cite{lime} or to guess model predictions only based on rationales~\cite{lertvittayakumjorn-toni-2019-human,shen2020useful}.
\section{Conclusion}
\label{sec:conclusion}


To investigate if the shortest rationales are best understandable for humans, this work presents a self-explaining model, \system, 
that achieves comparable performance with current self-explaining baselines in terms of end-task performance and human annotation agreement.
We further use \system to generate rationales for human studies to examine how rationale length can affect human understanding.
Our results show that the shortest rationales are largely not the best for human understanding.
This would encourage a rethinking of rationale methods to find the right balance between brevity and sufficiency.

\section{Acknowledgment}
\label{sec:acknowledgment}

We thank Chieh-Yang Huang for helpful comments on the paper, Bhargavi Paranjape for technical discussion of methods, and the crowd workers for participating in this study.
We also thank the anonymous reviewers for their constructive feedback.

\section{Ethical Considerations}
\label{sec:ethical}

This work shows that the shortest rationales are often not the best for human understanding.
We thus advocate for studying how users interact with machine-generated rationales.
However, we are aware that using rationales to interpret model prediction could pose some risks for users.
Rationales omit a significant portion of the contents (in our case, 50\% to 90\% of the words in a movie review are omitted), which could convey information incorrectly or mislead users.
Furthermore, machine-learned rationales could encode some unwanted biases~\cite{chuang2021mitigating}.
We believe that such risks should be explicitly communicated with users in real-world applications.

\bibliography{anthology}
\bibliographystyle{acl_natbib}

\newpage

\appendix
\section{Appendix}
\label{sec:appendix}


\subsection{Model Details and Hyperparameters}
\label{appdx:models}

\subsubsection{Methodology Details}

\paragraph{Concrete Relaxation of Subset Sampling Process.}
Given the output logits of \emph{identifier}, we use Gumbel-softmax~\cite{gumbelsoftmax:2017:iclr} to generate a concrete distribution as $\mathbf{c}=[c_1,...c_n]\sim\text{Concrete}(\mathbf{idn}(\mathbf{x}))$, represented as a one-hot vector over $n$ features where the top important feature is 1.
We then sample this process $k$ times in order to sample top-k important features, where we obtain $k$ concrete distributions as $\{\mathbf{c}^1,...,\mathbf{c}^k\}$.
Next we define one $n$-dimensional random vector $\mathbf{m}$ to be the element-wise maximum of these $k$ concrete distributions along $n$ features, denoted as $\mathbf{m} = \max_j \{\mathbf{c}_i^j\}_{i=n}^{j=k}$.
Discarding the overlapping features to keep the rest, we then use $\mathbf{m}$ as the k-hop vector to approximately select the top-k important features over document $\mathbf{x}$.

\paragraph{Vector and sort regularization.}
We deploy a \emph{vector and sort} regularization on mask $\mathbf{m}$~\cite{fong:2019:iccv}, where we sort the output mask $m$ in a increasing order and minimize the $L_1$ norm between $m$ and a reference $\hat{m}$ consisting of $n-k$ zeros followed by $k$ ones.

\subsubsection{Model Training Details}

\paragraph{Training and inference.}
During training, we select the Adam optimizer with the learning rate at 2e-5 with no decay.
We set hyperparameters in Equation~\ref{eq:objective} and~\ref{eq:regularizer} as $\lambda=1e-4$, $v_1=0.5$ and $v_2=0.3$ and trained 6 epochs for all models. 
Furthermore, we train \system~on a set of sparsity levels as $k=\{10\%, 20\%, 30\%, 40\%, 50\%\}$ and choose models with optimal predictive performance on validation sets.

\subsubsection{Details of Self-Explaining Baselines}
\label{appdx:baselines}

We compare our method with state-of-the-art self-explaining baseline models.
%


\paragraph{Sparse-N (Minimization Norm).}
This method learns the short mask with minimal $L_0$ or $L_1$ norm~\cite{lei-etal-2016-rationalizing,bastings-etal-2019-interpretable}, which penalizes for the total number of selected words in the explanation.

\begin{equation}
\begin{array}{l}
\displaystyle \min \  \mathbb{E}_{\mathbf{z}\sim\mathbf{idn}(\mathbf{x})} \mathcal{L}(\mathbf{cls}(\mathbf{z}),y)
+ \lambda ||m|| \\
\small
\label{eq:objective}
\vspace{-2.5mm}
\end{array}
\end{equation}


\paragraph{Sparse-C (Controlled Norm Minimization).}
This method controls the mask sparsity through a tunable predefined sparsity level $\alpha$~\cite{chang2020invariant,jain-etal-2020-learning}. The mask is penalized as below as long as the sparsity level $\alpha$ is passed.

\begin{equation}
\begin{array}{l}
\small
\displaystyle \min \  \mathbb{E}_{\mathbf{z}\sim\mathbf{idn}(\mathbf{x})} \mathcal{L}(\mathbf{cls}(\mathbf{z}),y) + \lambda \max (0, \frac{||\mathbf{m}||}{N}-\alpha) \\
\small
\label{eq:objective}
\vspace{-2.5mm}
\end{array}
\end{equation}


where N is the input length and $||m||$ denotes mask penalty with $L_1$ norm.

\paragraph{Sparse IB (Controlled Sparsity with Information Bottleneck).}
This method introduces a prior probability of $\mathbf{z}$, which approximates the marginal $p(\mathbf{m})$ of mask distribution; and $p(\mathbf{m}|\mathbf{x})$ is the parametric posterior distribution over $\mathbf{m}$ conditioned on input $\mathbf{x}$~\cite{paranjape-etal-2020-information}.
The sparsity control is achieved via the information loss term, which reduces the KL divergence between the posterior distribution $p(\mathbf{m}|\mathbf{x})$ that depends on $\mathbf{x}$ and a prior distribution $r(\mathbf{m})$ that is independent of $\mathbf{x}$.

\begin{equation}
\begin{array}{l}
\small
\displaystyle \min \  \mathbb{E}_{\mathbf{z}\sim\mathbf{idn}(\mathbf{x})} \mathcal{L}(\mathbf{cls}(\mathbf{z}),y) + \lambda KL[p(\mathbf{m}|\mathbf{x}), r(\mathbf{m})] \\
\small
\label{eq:objective}
\vspace{-2.5mm}
\end{array}
\end{equation}

\subsection{Ablation Study on Model Components}
\label{appdx:ablation}

We provide an ablation study on the Movie dataset to evaluate each loss term’s influence on end-task prediction performance, including Precision, Recall, and F1 scores. The result is shown in Table~\ref{tab:ablation}.

\begin{table}[!ht]
\centering
\begin{tabular}{llll}
\Xhline{2\arrayrulewidth}
\multirow{2}{*}{\textbf{Setups}} & \multicolumn{3}{l}{\textbf{End-Task Prediction}}                                            \\ \cline{2-4} 
                        & \multicolumn{1}{c}{\textbf{Precision}} & \multicolumn{1}{c}{\textbf{Recall}} & \textbf{F1} \\ \hline
\textbf{No Sufficiency} & \multicolumn{1}{c}{0.25}               & \multicolumn{1}{c}{0.50}            & 0.34        \\ \hline
\textbf{No Continuity}  & \multicolumn{1}{c}{0.82}               & \multicolumn{1}{c}{0.81}            & 0.81        \\ \hline
\textbf{No Sparsity}    & \multicolumn{1}{c}{0.80}               & \multicolumn{1}{c}{0.79}            & 0.79        \\ \hline
\textbf{No Contextual}  & \multicolumn{1}{c}{0.83}               & \multicolumn{1}{c}{0.83}            & 0.83        \\ \hline
\textbf{Our Model}      & \multicolumn{1}{c}{\textbf{0.91}}               & \multicolumn{1}{c}{\textbf{0.90}}            & \textbf{0.90}        \\ 
\Xhline{2\arrayrulewidth}
\end{tabular}
\caption{Ablation study of each module in our model on Movie Review dataset.}
\label{tab:ablation}
\end{table}

\subsection{Additional Details of Human Study}

\subsubsection{Generating Random Baselines}
\label{appdx:random_baseline}
%
%
%
%
Human accuracy likely increases when participants can see more words, {\em i.e.}, when the lengths of rationales increase.
If a rationale and a random text span have the same number of words, the rationale should help readers predict the label better.
We created a simple baseline that generated rationales by randomly selecting words to form the rationales.
We could control (1) how many words to select and (2) how many disjointed rationales to produce.
In the study, we set these two numbers to be identical to that of \system at each length level.

In detail, given the rationale length $k$, we first got the count of total tokens in rationale as \#tokens =
$k$.
Next, we computed the average number of rationale segments $m$, which are generated by \system, over the Movie dataset.
We randomly selected $m$ spans with total tokens' count as \#tokens from the full input texts, thus obtaining the random baselines.
We evenly separated 10 worker groups to finish five random baseline HITs and \system HITs each.
We determined that good model rationales should get higher human accuracy compared with same-length random baselines.

\subsubsection{Human Evaluation User Interface}
\label{appdx:interface}

\begin{figure*}[!t]
    \centering
    \includegraphics[width=.9\textwidth]{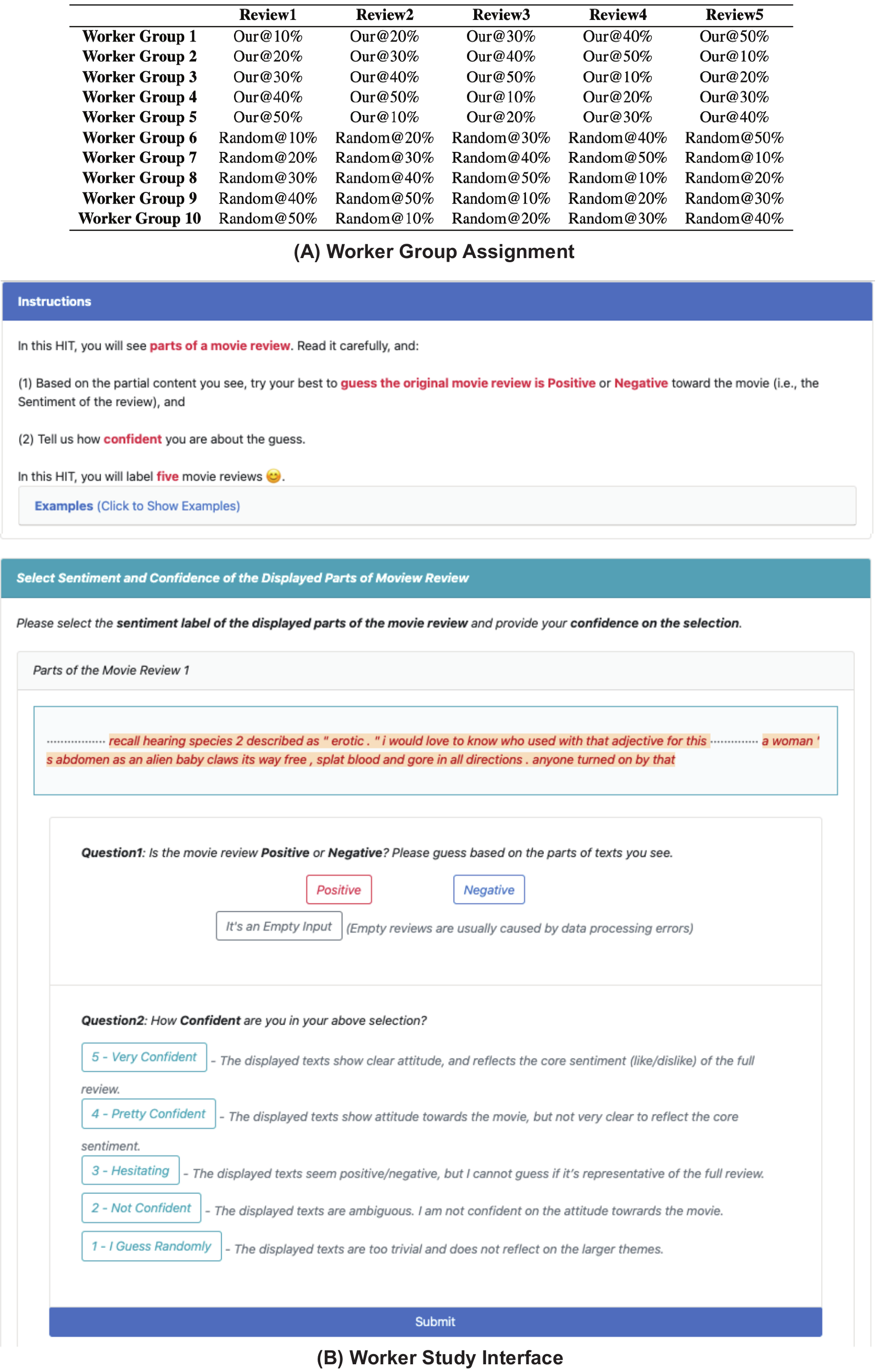}
    \caption{(A) The design of the worker group assignment in our human study. (B) The worker interface of the human study.
    }
    \label{fig:interface}
    \vspace{-4mm}
 \end{figure*}

\begin{figure*}[!t]
    \centering
    \includegraphics[width=0.95\textwidth]{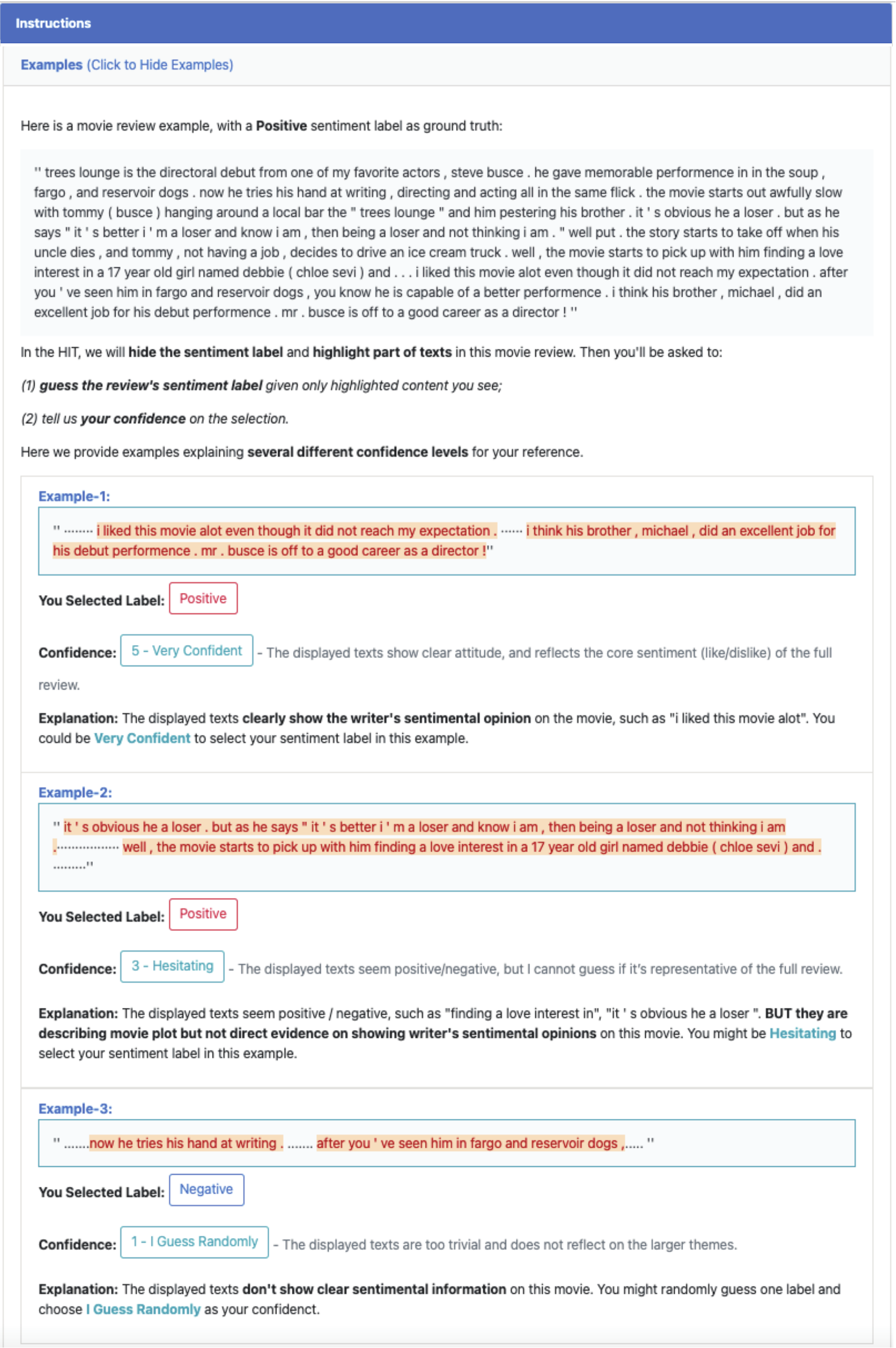}
    \caption{User Interface of the instruction in the human study.
    }
    \label{fig:instruction}
    \vspace{-4mm}
 \end{figure*}

We provide our designed user interfaces used in the human study. Specifically, we show the interface of the human study panel in Figure~\ref{fig:interface} (B). We also provide the detailed instructions for workers to understand our task, the instruction inteface is shown in Figure~\ref{fig:instruction}.


\end{document}